\documentclass[letterpaper]{article} 
\usepackage{aaai2026}  
\usepackage{times}  
\usepackage{helvet}  
\usepackage{courier}  
\usepackage[hyphens]{url}  
\usepackage{graphicx} 
\usepackage{natbib}  
\usepackage{caption} 
\usepackage{placeins}
\usepackage{algorithm}
\usepackage{algorithmic}
\usepackage{amsmath}
\usepackage{amssymb}
\usepackage{booktabs}
\usepackage{multirow}
\usepackage{subfig} 
\usepackage{algorithm}
\usepackage{algorithmic}
\usepackage{booktabs}
\usepackage{amssymb}
\usepackage{xcolor}
\usepackage{pifont}
\usepackage[subrefformat=parens]{subcaption} 
\usepackage{newfloat}
\usepackage{listings}
\usepackage{float}
\usepackage{makecell}
\usepackage{graphicx}
\usepackage{xcolor}
\usepackage{array}
\usepackage{rotating}
\usepackage{enumitem}
\usepackage{siunitx}  
\newcolumntype{C}[1]{>{\centering\arraybackslash}p{#1}}
\definecolor{lightgray}{gray}{0.92}
\usepackage[table]{xcolor}
\definecolor{lightgray}{RGB}{240,240,240}
\definecolor{highlight}{RGB}{255,255,200}
\usepackage{arydshln} 
\newcommand{\myparagraph}[1]{\noindent\textbf{#1}}
\DeclareCaptionStyle{ruled}{labelfont=normalfont,labelsep=colon,strut=off} 
\floatstyle{ruled}
\newfloat{listing}{tb}{lst}{}
\floatname{listing}{Listing}
\nocopyright 

\title{Self-Indexing KVCache: Predicting Sparse Attention  from Compressed Keys}
\author{
    Xu Yang\textsuperscript{\rm 1}\equalcontrib,
    Jiapeng Zhang\textsuperscript{\rm 1, \rm 2}\equalcontrib,
    Dongyang Zhao\textsuperscript{\rm 1},
    Guo Chen\textsuperscript{\rm 1},
    Zhuo Tang\textsuperscript{\rm 1, \rm3}\thanks{Zhuo Tang is the corresponding author.}
}
\affiliations{
    \textsuperscript{\rm 1}College of Computer Science and Electronic Engineering, Hunan University, Changsha, China\\
    \textsuperscript{\rm 2} The Ministry of Education Key Laboratory of “Fusion Computing of Supercomputing and Artificial Intelligence”  China\\
    \textsuperscript{\rm 3}Shenzhen Research Institute, Hunan University, Shenzhen, China\\

    {yangxv, zhangjp, ztang}@hnu.edu.cn

}

\usepackage{bibentry}

\begin{document}

\maketitle

\begin{abstract}
The KV cache in self-attention has emerged as a major bottleneck in long-context and large-batch inference for LLMs. Existing approaches often treat sparsity prediction and compression as separate modules—relying on auxiliary index structures to select relevant tokens, and on complex quantization schemes to reduce memory usage. This fragmented design introduces redundant overhead and limits scalability. 

In this paper, we propose a novel paradigm: treating the compressed key representation not merely as storage, but as a self-indexing structure that directly enables efficient sparse attention. By designing a sign-based 1-bit vector quantization (VQ) scheme, our method unifies compression and retrieval in a single, hardware-friendly format. This approach eliminates the need for external indices or learning-based predictors, offering a lightweight yet robust solution for memory-constrained inference. 
All components are designed to be hardware-efficient and easy to implement. By implementing  custom CUDA kernels, our method integrates seamlessly with FlashAttention, minimizing additional runtime and memory overhead. Experimental results demonstrate that our approach delivers both effectiveness and efficiency. \\

\end{abstract}
\begin{links}
    \link{Code}{https://github.com/LfieLike/selfindexingkv}
\end{links}
\section{Introduction}

Transformer-based large language models (LLMs)~\cite{dubey2024llama} have driven significant progress in NLP, achieving breakthroughs in numerous fields~\cite{yang2023humanintheloopmachinetranslationlarge,liu2023lostmiddlelanguagemodels,chen2021evaluatinglargelanguagemodels}. 
Self-attention~\cite{c:22}, as the core mechanism of Transformer, uses KV cache that consumes large amounts of memory and grows linearly with context length. This makes it a major bottleneck in large-batch inference and long-text inference.

Many efficient KV cache management strategies and optimization techniques have been proposed to ensure that the models can handle long contexts without exceeding resource limits.
LLMs have adopted techniques like Group-Query Attention (GQA)~\cite{ainslie2023gqa}, which enables multiple attention heads to share and process the same KV cache, reducing memory by several times. However, despite these advances, efficient KV cache management remains a key challenge in long-context inference scenarios.

To further alleviate the KV cache bottleneck, solutions such as sparsification, quantization, and low-rank factorization have been developed. However, optimizing the KV cache must balance three critical aspects: memory usage, latency, and inference accuracy. Each of the above approaches presents its own trade-offs in this balance. For example, static sparsification~\cite{NEURIPS2024_snapkv,zhang2023h2oheavyhitteroracleefficient} can significantly reduce memory usage and computational overhead but leads to noticeable drops in inference accuracy on certain tasks. 
Dynamic sparsification~\cite{tang2024quest,yang2025attentionpredictor} introduces non-negligible memory overhead due to the need for constructing token-wise indices that guide top-k selection during decoding. These indices improve relevance but increase memory usage, thereby affecting the overall memory-latency balance. Quantization~\cite{liu2024kivi,hooper2024kvquant10millioncontext,kang2024gear} and Low-rank factorization face inherent trade-offs between latency and memory reduction, and often suffer from format-specific constraints that limit optimization flexibility. In addition, some approaches rely on specialized data layouts or operators that are not readily supported by mainstream hardware, which may limit their practical deployment efficiency.

Additionally, learning-based methods~\cite{yang2025attentionpredictor, mazare2025inference} have been proposed, but their effectiveness is often tightly coupled with the distribution of the training data. This strong data dependency limits their generalizability across diverse tasks and scenarios.  These intertwined challenges underscore the current bottlenecks in KV cache management for LLM inference and call for more unified and principled solutions. To this end, an effective KV cache optimization method must satisfy three key requirements: 
\textbf{(i) Unbiased}. The optimization algorithm should generalize well across a wide range of downstream tasks. However, learning-based methods~\cite{yang2025attentionpredictor} rely heavily on auxiliary data, which compromises generalization. 
\textbf{(ii) Hardware-friendly}. Algorithms must be designed with hardware constraints in mind. An improperly designed complex algorithm can incur overheads that outweigh its intended benefits.
\textbf{(iii) Low Overhead}. Beyond theoretical efficiency, practical deployment requires minimal additional memory and compute overhead. However, many methods reduce one type of cost but increase others, and combining such techniques often leads to accumulated or conflicting overheads that negate the expected benefits.

\begin{figure}
    \centering
    \includegraphics[width=1\linewidth]{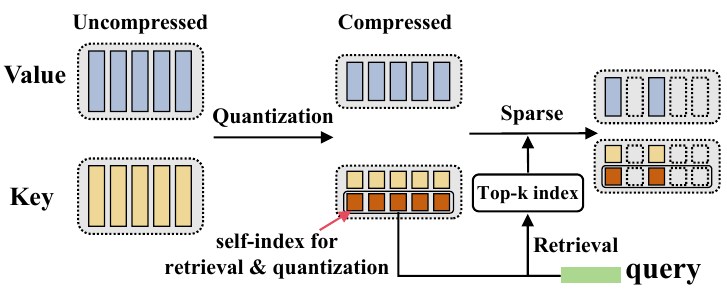}
    \caption{The core idea of our method: Find a \textbf{hardware-friendly} compressed representation for the key. This compressed data is designed to split into two parts, one of which enables fast retrieval with accuracy comparable to the full-precision key.}
    \label{fig:core_idea}
\end{figure}

In summary, these observations prompt a natural question: \textbf{Is it possible to develop a practical joint optimization strategy that preserves the complementary advantages of multiple techniques while minimizing redundant overhead}? To this end, as illustrated in Figure~\ref{fig:core_idea}, we design a unified optimization paradigm that integrates compression and sparsity in a mutually compatible manner. 
Unlike prior work~\cite{yang2025lserve}, which decouples compression and retrieval, our method co-designs both components and uses a single compact format that supports top-k selection directly in the compressed domain, thereby fully leveraging the existing information while avoiding redundant indexing or metadata overhead.

To instantiate this unified paradigm, we propose the Self-Indexing KVCache, a novel method that jointly optimizes dynamic sparsity and quantization. The key insight is that sparsity prediction in attention is essentially a top-k retrieval problem based on cosine similarity. To this end, we introduce a 1-bit vector quantization (VQ)-based retrieval algorithm, which allows fast and accurate token selection directly in the compressed domain. This retrieval process is accelerated by a custom CUDA kernel, ensuring low latency.
Meanwhile, our compression format is designed to support token-wise random access, making it fully compatible with attention acceleration frameworks such as FlashAttention~\cite{dao2022flashattention}. Crucially, the same 1-bit VQ indices used for retrieval also serve as the quantization representation, enabling efficient reuse and significantly reducing quantization error. By integrating sparsity prediction and quantization into a unified design, 
Self-Indexing KVCache achieves memory savings, speed improvements, and precision preservation simultaneously. It significantly outperforms the trade-offs of existing methods that optimize only a single dimension, thus realizing an all-round enhancement in both efficiency and effectiveness.
Our contributions are summarized as follows:
\begin{itemize} 
\item 
We propose a novel KV cache optimization paradigm that directly leverages compressed key cache for retrieval of the most valuable KV caches, thereby avoiding the indexing overhead inherent in dynamic sparsification.
\item 
We design a novel one-pass sign-bit-based VQ clustering strategy for retrieval index construction, enabling fast and expressive codebooks without iterative optimization.
\item Our method leverages custom-designed LUT-GEMV and sparse FlashAttention CUDA kernels for hardware-friendly optimization, minimizing computational overhead and memory traffic, achieving up to a 5× reduction in KV cache memory, 6.7× acceleration in sparse attention computation, and 2× speedup in end-to-end inference latency over FlashAttention v2.

\end{itemize}
\section{Background and Preliminaries}


\myparagraph{Self-attention and KV Cache.}
Self-attention is the core component of Transformer-based LLMs, formally expressed as $softmax\left( \frac{QK^T}{\sqrt{d}} \right)V$. During inference, the attention module undergoes two distinct stages: the prefill stage and the decode stage. 
Taking a single attention head as an example, in the prefill stage, the dimensions of Q, K, and V are ${L \times D}$, where $L$ is the context token length. As the context gets long, this stage is usually compute-intensive and fully utilizes GPU computational power. The K and V tensors are then cached as KV cache for computations in the decode stage. 
In the decode stage, the dimension of each Q becomes $1 \times D$. Since it requires storing and processing large context windows of tokens while generating predictions sequentially, this stage is memory-bound. 
The computational overhead can account for over 80\% of the total generation time. Therefore, optimizing the KV cache's storage and computational overhead during the decode stage is a critical research and engineering challenge.

\myparagraph{KV Cache Compression and Sparsity.}
Quantization is a standard technique for reducing KV cache memory, with KIVI~\cite{liu2024kivi} achieving 2-bit compression while preserving accuracy. More aggressive methods~\cite{zandieh2024qjl,NEURIPS2024_1bitVQ,kang2024gear,xiao2024smoothquantaccurateefficientposttraining,liu2024cachegen} achieve higher ratios but introduce non-trivial overhead. In parallel, sparsity-based approaches reduce compute by selecting a token subset. Static sparsity~\cite{han2024lminfinitezeroshotextremelength,xiao2024efficientstreaminglanguagemodels,NEURIPS2024_snapkv} often suffers from degraded accuracy, while dynamic methods~\cite{tang2024quest,liu2024clusterkv,xiao2024infllmtrainingfreelongcontextextrapolation} improve adaptivity via top-$k$ selection but rely on additional indexing structures.
Recent work explores tighter integration between compression and retrieval. 
Lserve~\cite{yang2025lserve} integrates sparse attention with low-bit KV cache compression, but treats them as separate components. Without joint optimization, this design introduces redundant overhead during inference. 
In contrast, our method directly reuses 1-bit quantized sign codes both for retrieval and reconstruction, eliminating redundancy between indexing and compression. This enables compressed-domain top-$k$ selection in a unified and hardware-efficient format, avoiding auxiliary predictors or metadata.

\begin{figure*}[t]
    \centering
    \includegraphics[width=0.9\linewidth]{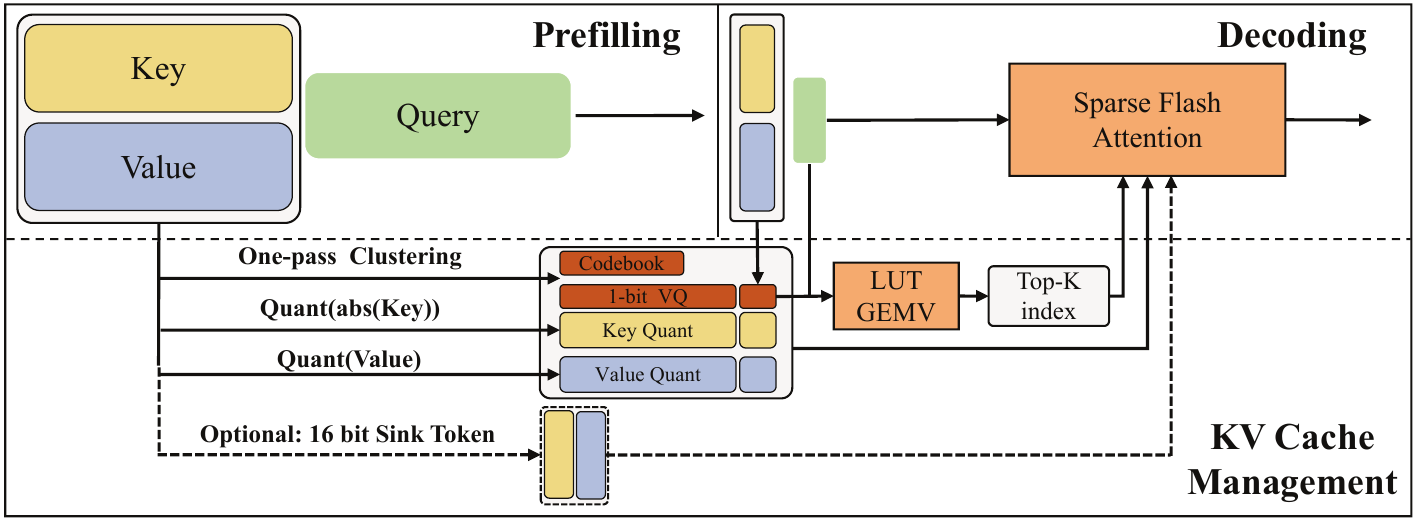}
        \caption{Overview of Self-Indexing KVCache. By leveraging  custom kernels, we significantly reduce the runtime overhead typically associated with retrieval and dequantization. Under low-bit quantization, we optionally preserve a fixed number of full-precision sink tokens during the prefill stage to improve inference robustness, \textbf{without sacrificing efficiency}.  }
    \label{fig:framework}
\end{figure*}

\label{sec:Motivation}  
\label{sec:Motivation3}  
\section{Implementation}

In this section, we focus on the overall algorithmic workflow of our proposed Self-Indexing KVCache. 
Figure~\ref{fig:framework} illustrates the Self-Indexing KVCache workflow. 
During the prefill stage, we perform one-pass sign-based vector quantization to construct a lightweight codebook and quantize the key and value tensors. 
This design avoids iterative clustering. As a result, it introduces minimal overhead and is well-suited for latency-sensitive applications. Optionally, 64 sink tokens can be retained in full precision. These tokens, along with the sink tokens, participate in the attention computation. By fusing dequantization into the flash attention kernel through a custom implementation, we significantly reduce overhead. 
Implementation details, including kernel-level optimizations, are deferred to the source code.
\subsection{One-Pass Sign-Based Clustering}
Sparsity prediction in attention can be formulated as a top-$k$ retrieval problem based on cosine similarity. To accelerate retrieval, VQ is commonly used to compress key vectors~\cite{zhang2024pqcacheproductquantizationbasedkvcache,liu2024clusterkv}. A key component of VQ is the construction of a codebook that clusters high-dimensional vectors into representative centroids.
Traditional codebook construction methods, such as K-means clustering, typically involve multiple iterations and introduce non-trivial computational overhead. While effective in some offline settings, their iterative nature can introduce considerable overhead during the prefill stage of LLM inference, where efficiency is critical.

To address this, we propose a one-pass, sign-based clustering method for efficient codebook generation. Instead of iteratively optimizing a reconstruction loss, we group vectors purely on the basis of their sign patterns. \textbf{In cosine similarity space, the sign of a vector conveys meaningful information about its direction}. Thus, clustering by sign patterns preserves the angular structure while drastically reducing computational overhead.
Experimental results demonstrate that this simple yet effective strategy enables the construction of a lightweight and expressive codebook in a single pass, making it well-suited for compressed-domain sparse attention under stringent inference-time constraints.

\myparagraph{Codebook Construction.} 
We begin by constructing a lightweight codebook based on sign patterns.  We partition the  key cache \( K \in \mathbb{R}^{L \times D} \) along the last dimension into \( G = D / 4 \) groups, each consisting of 4-dimensional subvectors:
\begin{equation}
K = [K^{(1)}, K^{(2)}, \dots, K^{(G)}], \quad K^{(g)} \in \mathbb{R}^{L \times 4}.
\end{equation}
Each subvector \( k \in \mathbb{R}^4 \) is encoded using its sign pattern:
\begin{equation}
s = \text{sign}(k) \in \{-1, +1\}^4.
\end{equation}
We define a deterministic mapping function \( \text{Code}(k): \mathbb{R}^4 \rightarrow \{0, \dots, 15\} \) that converts the sign pattern \( s \) into a 4-bit binary index by mapping \( +1 \rightarrow 1 \) and \( -1 \rightarrow 0 \). The resulting integer code is computed as:
\begin{equation}
\text{Code}(k) = \sum_{i=1}^{4} \left( \frac{1 + s_i}{2} \right) \cdot 2^{4 - i}.
\end{equation}
This assigns each subvector to one of the 16 sign-defined clusters. For each cluster, we compute a centroid by averaging the vectors assigned to it:
\begin{equation}
c_j = \frac{1}{|C_j|} \sum_{k \in C_j} k,
\end{equation}
where \( C_j \) denotes the set of subvectors sharing the same sign pattern. This process yields a compact codebook of size 16 for each group. The choice of 16 is a hardware-aware optimization: \textbf{a larger codebook would exceed the limited capacity of on-chip shared memory in modern accelerators, leading to higher latency and lower throughput.}
The entire procedure requires only a single pass and avoids costly iterative methods such as K-means~\cite{zhang2024pqcacheproductquantizationbasedkvcache}, resulting in minimal overhead while preserving the directional information essential for cosine-based sparsity prediction.

\myparagraph{Entropy-Aware Normalization.} 
While sign-based clustering enables efficient 1-bit VQ, applying it directly on raw key cache often results in unbalanced sign distributions, limiting the expressiveness of the resulting codes. To address this, we propose an entropy-aware normalization scheme that enhances the capacity of 1-bit representations from an information-theoretic perspective~\cite{shannon1948mathematical}.

Specifically, we apply channel-wise mean normalization to the original key cache \( K \in \mathbb{R}^{L \times D} \), subtracting the mean of each channel across all tokens:
\begin{equation}
K'_{i,d} = K_{i,d} - \mu_d, \quad \mu_d = \frac{1}{L} \sum_{i=1}^L K_{i,d}.
\end{equation}
This transformation ensures that each dimension has zero mean, which leads to a more balanced distribution of positive and negative signs.

From an information theoretic viewpoint, a binary variable \( B = \text{sign}(U) \in \{-1, +1\} \) achieves its maximum entropy when the probabilities of both outcomes are equal:
\begin{equation}
P(B = +1) = P(B = -1) = \frac{1}{2}, \quad \Rightarrow \mathcal{H}(B) = \log 2.
\end{equation}
Therefore, normalization increases the entropy of the sign patterns, improving the effective capacity of the 1-bit codes.

Importantly, this preprocessing does not affect the output of the attention mechanism, as the softmax function is invariant to additive shifts:
\begin{equation}
\text{softmax}(x - \mu) = \text{softmax}(x), \quad \forall \mu \in \mathbb{R}.
\end{equation}
Thus, our normalization improves quantization robustness without altering attention semantics.

\begin{figure}[t]
    \centering
    \includegraphics[width=0.95\linewidth]{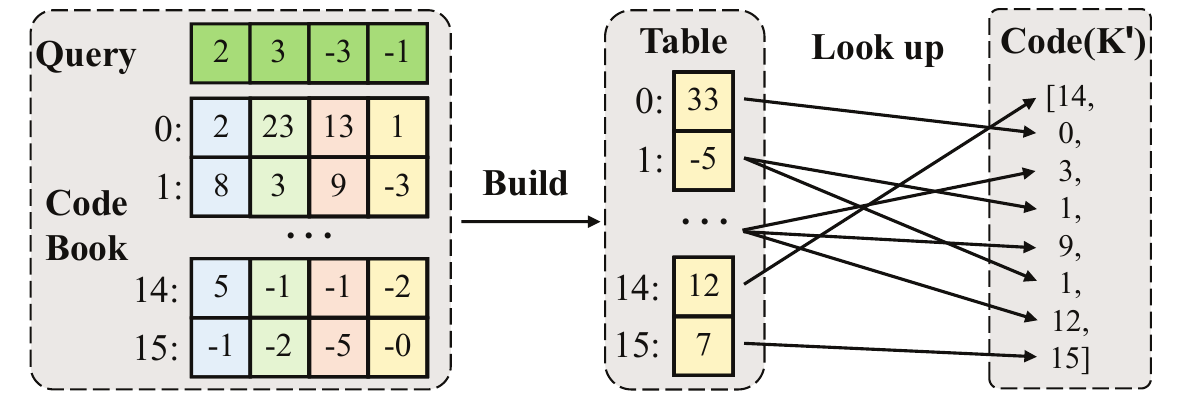}
    \caption{Overview of the LUT-GEMV. We compute the dot product between the query and each codeword in the codebook to generate a lookup table of size 16.}
    \label{fig:lookuptable1}
\end{figure}

\subsection{Compressed-Domain Top-$k$ Retrieval}

To enable efficient sparse attention, we perform similarity search between queries and the compressed key cache entirely in the quantized domain. As illustrated in Figure~\ref{fig:lookuptable1}, this process avoids full-precision computation and instead relies on fast table lookups.

Each query vector is first split into 4-dimensional subvectors, consistent with the group-wise structure used during key quantization. Each subvector is performed dot-produced with the 16 centroids in the codebook to compute similarity scores. These scores are precomputed and stored in a lookup table, where the index corresponds to one of the $2^4 = 16$ possible sign patterns per group.

The key cache $K$, already compressed into sign-pattern codes, stores only the index values representing which centroid each key subvector corresponds to. During retrieval, similarity scores between the query and all cached keys are efficiently approximated by performing simple table lookups followed by additions:
\begin{equation}
\text{score}(q, K) = qK^{T}  \approx \sum_{g=1}^{G} \text{Table}^{(g)}[\text{Code}(K')^{(g)}],
\end{equation}
where \( \text{Table}^{(g)}[\cdot] \) denotes the group-$g$ lookup table, and \( \text{Code}(K')^{(g)} \) is the index value of key \( k' \) in that group. 
The aggregated scores are then used to select the top-$k$ most relevant keys for sparse attention. 

Overall, this LUT-GEMV based method is hardware-efficient, easily parallelizable on GPUs. By replacing full-precision computations with lightweight table lookups and additions, our approach significantly reduces floating-point operations and memory bandwidth usage—two major bottlenecks in large-scale inference. Despite this simplification, it maintains high retrieval accuracy, demonstrating the effectiveness of compressed-domain similarity computation.

\begin{table*}[ht]

\centering

{%
\setlength{\tabcolsep}{2pt}
\renewcommand\arraystretch{0.3}
\begin{tabular}{@{}l ccccccccccccccc@{}}
\toprule
\multirow{2}{*}[-1pt]{\makecell[l]{\hspace{10pt}\raisebox{0pt}{Method}}} & \multirow{2}{*}[1pt]{\makecell[c]{Cache Bits\\(K,V,Index)}} & \multicolumn{2}{c}{SD-QA} & \multicolumn{2}{c}{MD-QA} & \multicolumn{2}{c}{Summarization} & \multicolumn{2}{c}{Few-shot} &\multicolumn{2}{c}{Synthetic}
& \multicolumn{2}{c}{Code} & \multirow{2}{*}[-1pt]{Avg.} \\ 
\cmidrule(lr){3-4} \cmidrule(lr){5-6} \cmidrule(lr){7-8} \cmidrule(lr){9-10} \cmidrule(lr){11-12} \cmidrule(lr){13-14} 
& & \rotatebox{1}{Qasper} & \rotatebox{1}{MF-en} & \rotatebox{1}{HPQA} & \rotatebox{1}{2WQA} & \rotatebox{1}{GVRpt} & \rotatebox{1}{QMSum} & \rotatebox{1}{TREC} & \rotatebox{1}{TrivQA} &
\multicolumn{2}{c}{\rotatebox{1}{PR-en}} & \rotatebox{1}{Lcc} & \rotatebox{1}{RB-P} &  \\
\midrule





\rowcolor{lightgray}
Llama3.1-8B & 16, 16, 0& 45.5 & 53.8 & 54.7 & 47.1 & 34.9 & 25.3 & 73.0 & 91.6 & \multicolumn{2}{c}{99.5} & 63.4 & 56.7 & 58.7  \\

\midrule
SnapKV & 16, 16, 0  & 33.9 & 46.4 & 54.4 & 44.7 & 21.8 & 22.6 & 48.0 & 90.5 & \multicolumn{2}{c}{78.0} & 58.0 & 50.1 & 49.9  \\

Quest & 16, 16, 2 & 44.9 & 51.6 & 55.0 & 46.9 & 31.2 & 24.0 & 71.0 & 90.6 & \multicolumn{2}{c}{95.5} & 60.6 & 50.5 & 56.5  \\

DoubleSparse & 16, 16, 2 &  43.8 & 51.7 & 54.4 & 45.9 & 31.7 & 24.0 & 67.5 & 91.9 & \multicolumn{2}{c}{97.0} & 50.5 & 44.7 & 54.8  \\

Ours(16 bits) & 16, 16, 1& \underline{45.4} & {53.9} & \underline{55.4} & \underline{47.0} & \underline{34.0} & \underline{24.8} & \underline{72.5} & {91.7} & \multicolumn{2}{c}{\underline{99.5}} & \underline{63.2} & \underline{55.3} & \underline{58.4}  \\


Ours & 2, 2, 1 & 44.8 & \underline{54.6} & 55.7 & 46.0 & 33.1 & 24.7 & 72.5 & \underline{92.1} & \multicolumn{2}{c}{ 99.5} & 63.2 & 54.4 & 58.2 \\
\midrule
\rowcolor{lightgray}
Qwen2.5-14B & 16, 16, 0 & 46.5 & 54.0 & 65.0 & 64.3 & 31.9 & 23.6 & 81.0 & 89.0 & \multicolumn{2}{c}{100.0} & 33.1 & 38.1 & 56.9 \\

\midrule
SnapKV & 16, 16, 0 & 29.4 & 47.6 & 60.1 & 54.0 & 19.3 & 19.8 & 54.5 & 86.0 & \multicolumn{2}{c}{61.0} & 24.9 & 31.1 & 44.3  \\

Quest & 16, 16, 2 & 45.7 & 52.0 & 62.6 & 62.3 & 28.8 & 21.3 & 70.5 & 88.1 & \multicolumn{2}{c}{94.5} & 31.7 & 35.2 & 53.9  \\

DoubleSparse & 16, 16, 2& 44.7 & 53.1 & 61.0 & 62.1 & 30.3 & 21.9 & 79.0 & \underline{89.0} & \multicolumn{2}{c}{94.1} & 27,5 & 31.8 & 54.1  \\

Ours(16 bits) & 16, 16, 1 & \underline{46.0} & \underline{53.5} & \underline{63.7} & 62.4 & \underline{30.5} & \underline{22.6} & \underline{80.5} & 88.5 & \multicolumn{2}{c}{\underline{97.5}} & \underline{32.6} & \underline{37.4} & \underline{55.9}  \\

Ours & 2, 2, 1  & 45.8 & 53.5 & 63.0& \underline{63.1} & 30.5 & 22.0 & 80.5 & \underline{88.5} & \multicolumn{2}{c}{96.4} & 32.3 & \underline{37.4} & 55.7  \\

\bottomrule
\end{tabular}%
}
\caption{Performance comparison  of LongBench. The best result is highlighted  in \underline{underline}.}
\label{tab:longbench_llama2}
\end{table*}

\subsection{Token-Wise Quantization Format}
\myparagraph{Token-Wise vs. Channel-Wise.} 
In random-access or token-level decoding scenarios, token-wise quantization offers clear efficiency advantages. Channel-wise quantization, as used in prior work~\cite{liu2024kivi}, stores quantization parameters per channel, requiring all dimensions to be read in order to reconstruct a single token, resulting in high latency and memory bandwidth. This design may be effective in dense settings, but becomes inefficient for sparse access. In contrast, token-wise quantization stores parameters per token, significantly reducing parameter lookup overhead, making it naturally suited for efficient sparse token retrieval.

\myparagraph{Quantization Function.}
Given an input value cache \( \mathbf{V} \), we compute the quantization scale \( qs \) and zero point \( zp \) as:
\begin{equation}
qs = \frac{{V}_{\max} - {V}_{\min}}{2^B - 1}, \quad zp = {V}_{\min}.
\end{equation}
The quantized values are obtained by:
\begin{equation}
Q({V}) = \text{clamp} \left( \text{round} \left( \frac{{V} - zp}{qs} \right), 0, 2^B - 1 \right).
\end{equation}

Dequantization is then performed as:
\begin{equation}
D(V) = qs \cdot Q(\mathbf{V}) + zp.
\end{equation}

Given a key cache matrix \( \mathbf{K} \), we first take the absolute value and normalize it across each channel (dimension) using the per-channel maximum:
\begin{equation}
\hat{\mathbf{K}} = \frac{|\mathbf{K}|}{\alpha}, \quad \text{where} \quad \alpha_j = \max \left( |\mathbf{K}_{:,j}| \right).
\end{equation}

As the sign information is already extracted, it is excluded from the subsequent quantization. Only the absolute values  are quantized and stored.

The per-channel scaling factors \( \alpha \) are also \textbf{reused during the decoding stage}. Finally, the dequantized key values are recovered as:
\begin{equation}
D(|K|) = \alpha \cdot (qs \cdot Q(|\hat{K}|) + zp).
\end{equation}
Experimental results demonstrate that our method enables effective 2-bit token-wise quantization without significant loss in performance.

\myparagraph{Full Precision Sink Tokens.}
We observe that certain tokens are consistently selected and are particularly sensitive to low-bit quantization. To address this, we adopt the SnapKV~\cite{NEURIPS2024_snapkv} in the prefill stage, where we fix the selection of 64 tokens in full precision. These tokens are guaranteed to always participate in the sparse attention computation, mitigating quantization-induced errors. 
This trick provides robustness without sacrificing efficiency: the overhead is negligible relative to total KV cache size, and the fixed, contiguous layout allows seamless integration into our FlashAttention kernel.  

\subsection{Overhead Analysis}
For memory overhead. 
Take the KV cache of an attention head in LlaMA-3.1-8B with a context length of \( L \), where \( \mathbf{K}, \mathbf{V} \in \mathbb{R}^{L \times 128} \). The memory required under our method includes the following components: \textbf{(i)} Sign bits: Each key vector requires \( L \times 128 \) bits for storing sign information. 
\textbf{(ii)} Quantized values and parameters: Both keys and values are 2-bit quantized, requiring \( 2 \times (2 \times L \times 128) = 512L \) bits. In addition, the quantization is applied per 32 elements, yielding \( 4L \) groups for each of key and value. Each group stores a 16-bit scale and zero point, leading to another \( 2 \times 4L \times 2 \times 16 = 256L \) bits. In total, this occupy \( 768L \) bits. 
\textbf{(iii)} Fixed overhead: The codebook, normalization parameters $\alpha$ and 
$\mu$, as well as memory for handling outlier tokens, contribute a fixed overhead. As 
$L$ increases, this overhead becomes negligible.
 Therefore, our method achieves up to 78\% memory savings for the KV cache without degrading inference accuracy.
Furthermore, We implement all components with custom CUDA kernels to minimize memory transfers and maximize runtime efficiency.




\section{Evaluation}

\subsection{Experimental Setup}
\myparagraph{Benchmark and Module.}
We evaluate our method using LongBench~\cite{bai2024longbenchbilingualmultitaskbenchmark}, which covers six categories: Single/Multi-Document QA, Summarization, Few-shot Learning, Synthetic Tasks, Code Completion. For each category, we select two representative datasets. Additionally, we use the Ruler benchmark~\cite{hsieh2024ruler} to assess ultra-long context performance.
We evaluate our method on two open-source models: Llama3.1-8B-Instruct~\cite{dubey2024llama} (128K context with GQA), and Qwen2.5-14B-1M~\cite{yang2024qwen2} (1M context with aggressive GQA). Through experiments on these models, we demonstrate the robustness of our approach.

\myparagraph{Baseline.}
We consider SnapKV~\cite{NEURIPS2024_snapkv}, Quest~\cite{tang2024quest}, and DoubleSparse~\cite{yang2024postdoublespars} as our baselines. We implement all methods to ensure fair comparison: SnapKV serves as a standard one-shot KV cache pruning baseline, known for its simplicity and efficiency. Quest employs dynamic block-wise sparse attention with effective prediction strategies. DoubleSparse offers fine-grained token-wise sparsity with low overhead. Together, these methods span a spectrum from static pruning to dynamic sparsity, forming a comprehensive comparison baseline.
Recent methods such as SAAP~\cite{mazare2025inference}  adopt fundamentally different designs, relying on auxiliary encoders, separate indices, or heavy offline preprocessing. These methodological gaps make direct and fair comparisons infeasible. Instead, we evaluate against open-sourced, lightweight baselines that align with our  inference setting.

\myparagraph{Hyperparameter Settings.}
To ensure a fair comparison, we meticulously align extraneous variables. For Quest, we set the chunk size to 16.
For DoubleSparse, we use 16 channels for token selection, which is equivalent to building a 2-bit index per parameter over the key cache. 
For all methods, sparse attention is applied at every layer, and tokens generated during the decode stage are always included in the attention computation by default. Experiments are conducted on NVIDIA RTX 4090 and A100-40G.

\subsection{Performance  Study}


\begin{table*}[!t]
\centering

\label{tab:64k_performance}
{%
\setlength{\tabcolsep}{3pt}
\renewcommand\arraystretch{1}
\begin{tabular}{@{}l cccccccccccccccc@{}}
\toprule
\multirow{2}{*}[-1pt]{\makecell[l]{\hspace{10pt}\raisebox{0pt}{Method}}} & 
\multirow{2}{*}[-1pt]{\makecell[c]{Cache Bits\\(K,V,Index)}} & \multicolumn{13}{c}{Task} & \multirow{2}{*}[-1pt]{\makecell[c]{Avg.}} \\ 
\cmidrule(lr){3-15} 
& & NS1 & NS2 & NS3 & NM1 & NM2 & NM3 & NV & NQ & VT & CWE& FWE & QA1 & QA2 &  \\
\midrule
\rowcolor{lightgray}
Llama3.1-8B & 16,16,0 & 100.0&100.0&100.0&100.0&100.0&100.0&100.0&100.0&98.4&59.2&94.67&72.0&56.0& 90.8\\
\midrule
SnapKV & 16,16,0 & 100.0 & 100.0 & 28.0 & 100.0 & 32.0 & 8.0 & 100.0 & 100.0 & 96.8 & 43.6 & 90.0&60.0  & 56.0 & 70.8  \\
Quest & 16,16,2 & 100.0 & 100.0 & 100.0 & 100.0 & 100.0 & 56.0 & 98.0 & 100.0 & 97.6 & 31.6 & 90.7 & 68.0&  56.0& 84.5 \\
DoubleSparse & 16,16,2 & 100.0 & 100.0 & 100.0 & 100.0 & 68.0 & 68.0 &100.0 & 100.0 & 97.6 & 34.8 & 90.6& 64.0 & 64.0 & 83.6  \\
Ours(16 bits) & 16,16,1 & 100.0 & 100.0 & 100.0 & 100.0 & 100.0 & 100.0 & 98.0 & 100.0 & 98.4 & 53.2&  89.3 & 68.0 & 60.0 & 89.4  \\
Ours & 2,2,1 & 100.0 & 100.0 & 100.0 & 100.0 & 100.0 & 96.0 & 100.0 & 100.0 & 96.8 & 49.2& 85.3 & 68.0 & 64.0 &  89.2 \\
\midrule
\rowcolor{lightgray}
Qwen2.5-14B & 16,16,0 & 100.0 & 100.0 & 100.0 & 100.0 & 100.0 & 100.0 & 99.0 & 100.0 & 100.0 & 90.8 & 94.7 & 80.0 & 72.0 & 95.1\\
\midrule

SnapKV          & 16,16,0 & 100.0 & 100.0 & 52.0  & 100.0 & 0.0   & 4.0   & 94.0 & 100.0 & 100.0 & 75.0 & 94.67 & 68.0 & 60.0 & 72.9 \\
Quest         & 16,16,2 & 100.0 & 100.0 & 100.0 & 100.0 & 100.0 & 96.0  & 99.0 & 100.0 & 100.0 & 76.4 & 90.67 & 80.0 & 76.0 & 93.7 \\
DoubleSparse            & 16,16,2 & 100.0 & 100.0 & 92.0  & 100.0 & 88.0  & 100.0 & 100.0 & 98.0  & 100.0 & 89.6 & 93.33 & 80.0 & 72.0 & 93.3 \\
Ours(16 bits) & 16,16,1 & 100.0 & 100.0 & 100.0 & 100.0 & 100.0 & 100.0 & 99.0 & 100.0 & 100.0 & 93.6 & 94.6 & 80.0 & 68.0 & 95.0 \\
Ours     & 2,2,1 & 100.0 & 100.0 & 100.0 & 100.0 & 100.0 & 100.0 & 98.0 & 99.0  & 100.0 & 83.6 & 92.0  & 80.0 & 68.0 & 93.9 \\

\bottomrule
\end{tabular}%
}
\caption{Detail experimental results of Llama3.1-8B on the 32K prompt Ruler Benchmark.}
\end{table*}

\myparagraph{LongBench.}
In the performance comparison experiment of LongBench with a budget of 160 tokens, our method retains the 64 sink tokens, thus only dynamically select 96 tokens.  We compared the performance of various methods on different tasks for different models. As shown in Table~\ref{tab:longbench_llama2}, under the same settings, our method achieves lower accuracy degradation. Even with 2-bit quantization, the accuracy loss remains minimal, thanks to the use of sign-bit assistance. 

\begin{figure}
    \centering
    \includegraphics[width=1\linewidth]{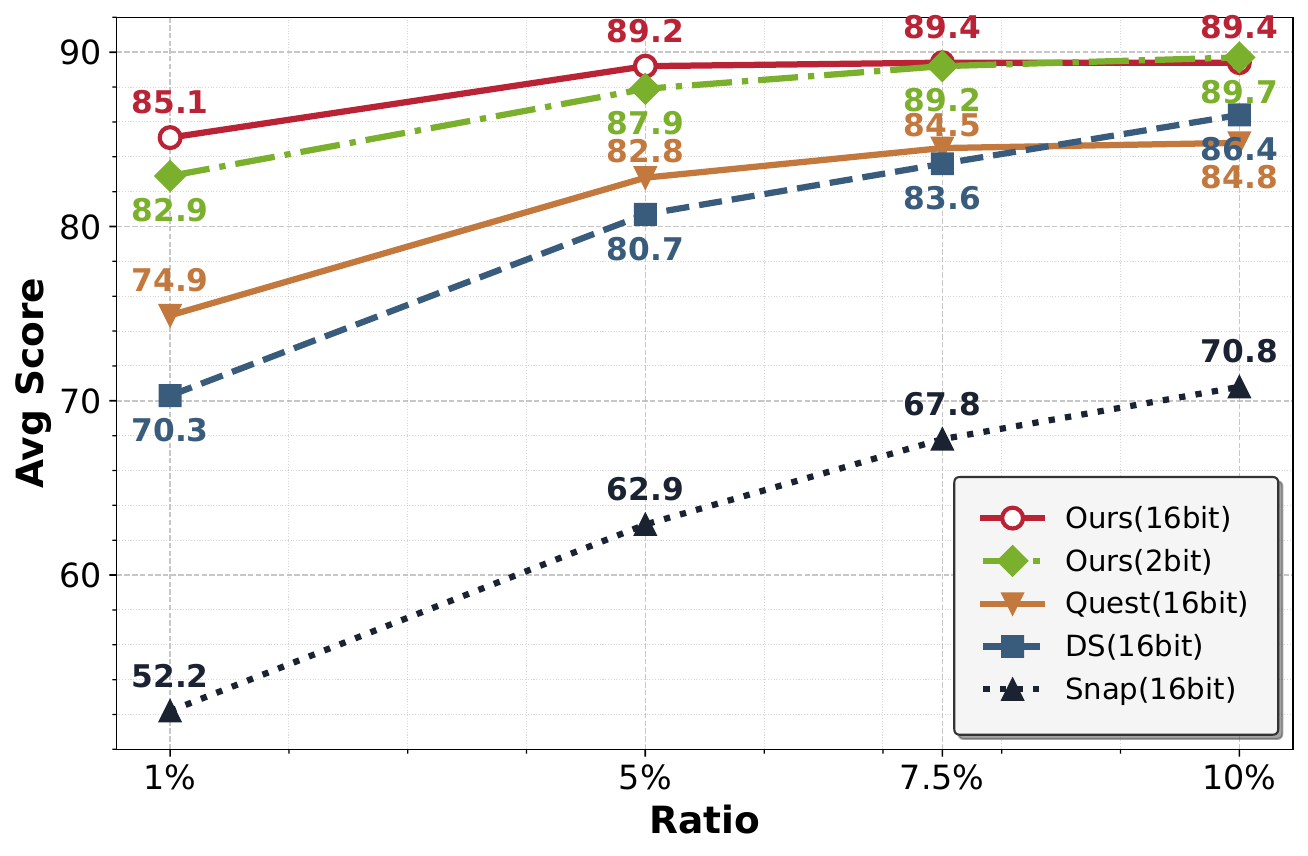}
    \caption{Average task performance under different sparsity ratios on the 32K prompt Ruler Benchmark. Our method outperforms other method, with the 2-bit quantization at 7.5\% sparsity.}
    \label{fig:rate}
\end{figure}

\myparagraph{RULER.}
In long-context experiments, we retain a proportion of tokens for attention computation rather than a fixed number. As shown in Figure~4, our method achieves optimal accuracy even with only 7.5\% of tokens preserved. Table~2 presents results across all tasks under this 7.5\% sparsity setting, focusing on 32K-token prompts. We choose this length as both models show noticeable degradation beyond 32K, making comparisons less meaningful. Even under such extreme sparsity, our method consistently outperforms all baselines. Notably, it shows more stable performance on reasoning tasks like FWE and CWE, suggesting that our dynamic retrieval better captures critical information.


\subsection{Efficiency Study}
\begin{table}[t]
\centering
\begin{tabular}{c|ccc}
\hline
\textbf{Prompt Length} & \textbf{Ours} & \textbf{KIVI} & \textbf{Flash Attention2} \\
\hline
8K  & 1.26   & 1.23   & 1.18 \\
16K & 2.75   & 2.66   & 2.62 \\
32K & 6.51   & 6.32   & 6.18 \\
48K & 11.12  & 10.88  & OOM \\
64K & 17.70  & OOM    & OOM \\
\hline
\end{tabular}
\caption{Comparison of Time TT2T (seconds) across different methods and prompt lengths.}
\label{tab:tt2t_comparison}
\end{table}


\begin{figure}[t]
    \centering
    \includegraphics[width=1\linewidth]{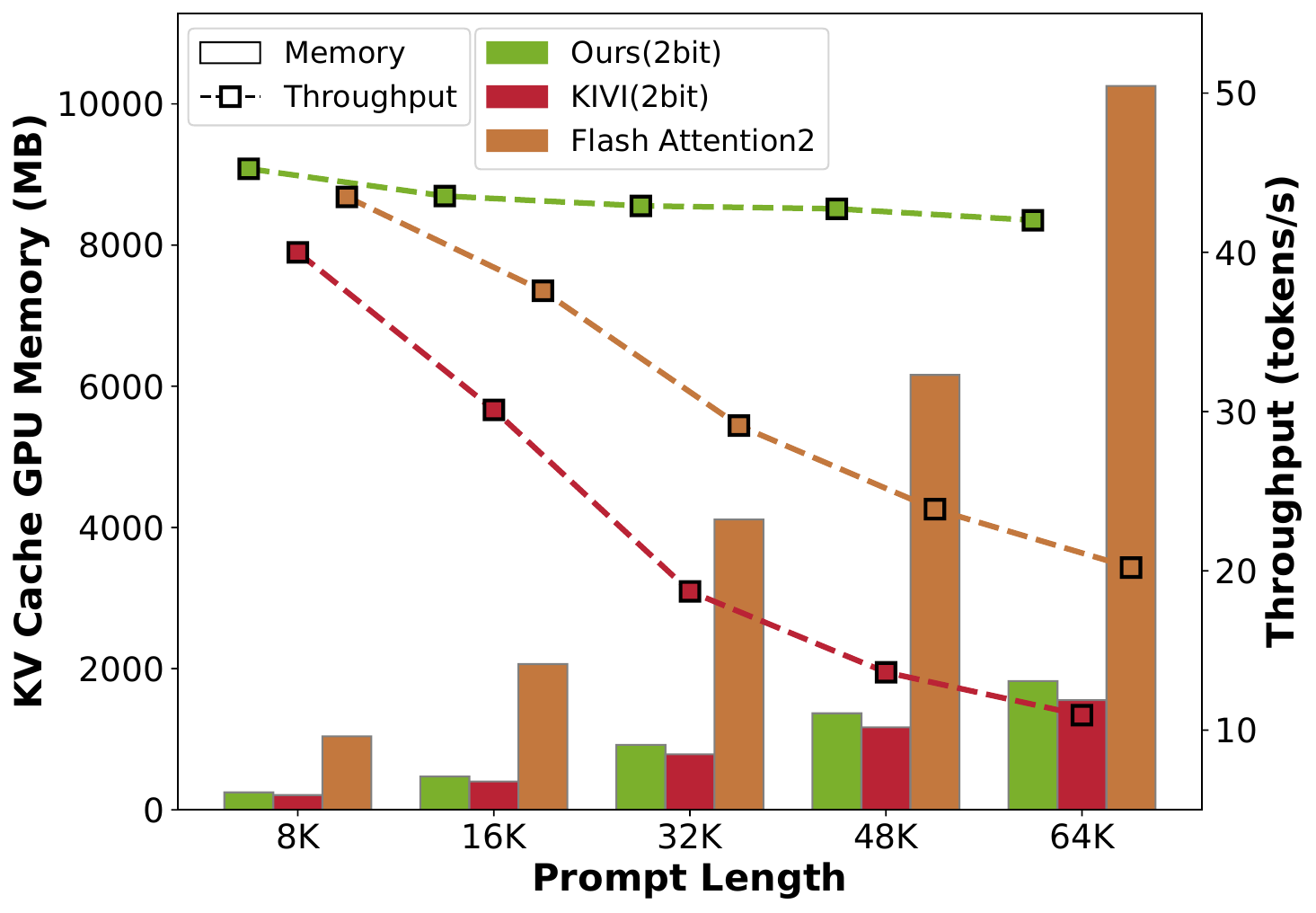}
    \caption{KV cache memory overhead and latency of LlaMA3.1-8B at different prompt lengths.  Our method dynamically select 7.5\% tokens. 
}
    \label{fig:latency_memory}
\end{figure}

\myparagraph{End to End Efficiency.}
We implemented the Self-index Cache based on the Transformers framework~\cite{wolf2019huggingface} and conducted end-to-end efficiency comparisons. To ensure fairness and strong baseline performance, we compared against full cache, Flash Attention2, and KIVI 2-bit quantization, all using the official implementations from the Transformers library. We adopt Time To 2nd Token (TT2T), as the evaluation metric in the prefill stage, and report KV cache memory footprint and throughput as metrics in the decode stage. Notably, our method retains only 7.5\% of the tokens in the cache.

Table~\ref{tab:tt2t_comparison} presents the TT2T evaluation results, from which we draw two key conclusions:
\textbf{(i)} Compared to Flash Attention2, our method introduces only 5\% additional overhead, indicating that both the quantization process and codebook construction are highly efficient and lightweight.
\textbf{(ii)} Compared to KIVI, our method supports significantly longer context lengths, attributed to our custom sparse Flash Attention kernel.We fuse dequantization and sparse memory access into a single compute pass, minimizing memory traffic and maximizing execution efficiency.
These results highlight not only the scalability of our approach under longer sequences, but also the importance of low-level kernel optimization in achieving high-performance sparse attention.

Figure 5 illustrates the performance of the three methods during the decode stage, leading to the following key observations:
Storage Efficiency: Our method achieves nearly 5× reduction in KV cache memory footprint, matching the compression ratio of 2-bit KIVI while maintaining retrieval compatibility.
Throughput Advantage: Thanks to aggressive sparsification and a fused kernel design, our approach delivers up to 2× higher decoding throughput compared to full-cache FlashAttention2. In contrast, KIVI suffers from degraded performance due to its naive decompress-then-compute strategy, underscoring the efficiency of our integrated implementation.

\myparagraph{Head to Head Efficiency.}
Table~\ref{tab:module_comparison} demonstrates that our method achieves strong efficiency across all major components.
\textbf{(i)} The proposed one-pass sign-based clustering achieves over 20× speedup compared to traditional KMeans while preserving sufficient representational quality for downstream retrieval. Here, we follow prior work on KV cache clustering, which typically adopts 20–50 KMeans iterations for convergence~\cite{zhang2024pqcacheproductquantizationbasedkvcache}.
\textbf{(ii)} Our retrieval procedure, which includes table construction and LUT-GEMV computation, provides more than 4× speedup over full dot-product attention and outperforms Quest under identical settings.
\textbf{(iii)} In the attention stage, our custom sparse FlashAttention kernel achieves a 6–7× speedup relative to full FlashAttention2, while maintaining comparable efficiency to Page Attention, which is commonly used in prior sparse attention methods such as Quest to accelerate block-level computation.

These results highlight the effectiveness of our design in minimizing latency at every stage of the inference pipeline.
\begin{table}[t]
\centering

\begin{tabular}{lll}
\toprule
Module & Method & Time (ms) \\
\midrule
\multirow{2}{*}{Clustering} 
    & Ours & 88.85  \\
    & KMeans (20 iterations) & 2113.9  \\
\midrule
\multirow{3}{*}{Retrieval} 
    & Ours & 0.039  \\
    & Quest (page size = 16) & 0.041  \\
    & Full $K \cdot q^\top$ & 0.166  \\
\midrule
\multirow{3}{*}{Attention} 
    & Ours (7.5\%) & 0.116 \\
    & Page Attention (7.5\%) & 0.101 \\
    & Flash Attention2 (Full) & 0.776 \\
\bottomrule
\end{tabular}
\caption{Performance comparison across different modules under a 16K token input with batch size 10.}
\label{tab:module_comparison}
\end{table}

\begin{table}[t]
\centering
\renewcommand{\arraystretch}{1.1} 

\begin{tabular}{lcccc}
\toprule
Setting & MF-en & HPQA & GovRpt  & RB-P \\
\midrule
Ours & \textbf{54.6} & \textbf{55.7} & \textbf{33.1}  & \textbf{54.4} \\
w/o sign in quant & 52.5 & 54.9 & 32.2 &  50.9 \\
sign-only retrieval & 52.8 & 53.2 & 31.9 &  52.1 \\
w/o sink tokens & 52.6 & 55.2 & 32.5 &  54.2 \\
\bottomrule
\end{tabular}
\caption{Ablation study on the effect of components.}
\label{tab:ablation_1bitvq}
\vspace{-3mm}
\end{table}

\subsection{Ablation Study}
 The results are shown in Table~\ref{tab:ablation_1bitvq}. To assess the contributions of key components in our design, we perform ablation experiments on sign-bit quantization, retrieval mechanism, and the use of sink tokens.
Removing the sign bit during quantization (w/o sign in quant) consistently reduces accuracy, confirming that sign information carries essential directional cues that mitigate low-bit quantization error.
The sign-only retrieval variant, which computes similarity using only the sign bit without magnitude-based VQ, results in even larger degradation. This indicates that the magnitude component plays a critical role in accurate token selection. Removing sink tokens has minor impact on most datasets but slightly reduces robustness in low-redundancy settings like GovRpt.
Overall, these results highlight the dual role of 1-bit VQ and the complementary effect of sink tokens in inference quality under aggressive compression.



\section{Conclusion}
In this work, we presented Self-Indexing KVCache, a practical and hardware-friendly method that unifies compression and sparse retrieval for attention in large language models. By using the compressed representation itself as a functional index, our design directly supports top-$k$ token selection in the attention computation process. Leveraging sign-bit-based 1-bit vector quantization, entropy-aware normalization, and a custom LUT-GEMV kernel, our method achieves up to 5$\times$ memory reduction and consistently lower latency, with negligible accuracy loss. Experiments demonstrate that Self-Indexing KVCache outperforms strong baselines in both long-context reasoning and end-to-end inference efficiency, all without requiring extra training or auxiliary indexing structures.
These findings suggest that compression in LLM systems can be more than a storage optimization; when co-designed with retrieval, it can act as a computation-aware, index-equivalent representation, opening new opportunities for unified and efficient inference in future large-scale models.
\section{Acknowledgments}
The work is supported by the National Natural Science Foundation of China (Grant Nos. 62225205, 62302160, 62222204), the Natural Science Foundation of Hunan Province (Grant Nos. 2024JJ6154), the Science and Technology Program of Changsha (kh2301011), the Major Science and Technology Research Projects of Hunan Province (Grant Nos. 2024QK2010, 2024QK2009), the Yunnan Provincial Major Science and Technology Special Plan Projects (No.202502AD080009), the Shenzhen Basic Research Project (Natural Science Foundation) under Grant JCYJ20210324140002006.

\bibliography{aaai2026}

\end{document}